\documentclass{article}
\usepackage{spconf,amsmath,graphicx,amsfonts}
\DeclareMathOperator*{\argmin}{arg\,min}

\title{Multi-scale Octave Convolutions for Robust Speech Recognition}
\name{Joanna Rownicka, Peter Bell, Steve Renals\thanks{This work was supported by a PhD studentship from the DataLab Innovation Centre, Ericsson Media Services, and Quorate Technology.}}
\address{Centre for Speech Technology Research, University of Edinburgh, UK}

\begin{document}
%\ninept
%
\maketitle
\begin{abstract}
%[DONE]
We propose a multi-scale octave convolution layer to learn robust speech representations efficiently.
%with a more efficient CNN design. 
Octave convolutions were introduced by Chen et al~\cite{OctConv} in the computer vision field to reduce the spatial redundancy of the feature maps by decomposing the output of a convolutional layer into feature maps at two different spatial resolutions, one octave apart.
This approach improved the efficiency as well as the accuracy of the CNN models. The accuracy gain was attributed to the enlargement of the receptive field in the original input space. We argue that octave convolutions likewise improve the robustness of learned representations due to the use of average pooling in the lower resolution group, acting as a low-pass filter. We test this hypothesis by evaluating on two noisy speech corpora -- Aurora-4 and AMI. We extend the octave convolution concept to multiple resolution groups and %we reduce the spatial resolution by using 
multiple octaves. 
%By doing this, we aim to further improve the robustness of learned representations while also reducing the computational cost. 
To 
%improve the explainability of our models and to 
evaluate the robustness of the inferred representations, we report the similarity between clean and noisy encodings using an affine projection loss as a proxy robustness measure. The results show that proposed method reduces the WER by up to 6.6\% relative for Aurora-4 and 3.6\% for AMI, while improving the computational efficiency of the CNN acoustic models.

\end{abstract}
%
% \begin{keywords}
% robust speech representations, CNNs, octave convolutions
% \end{keywords}
%
\section{Introduction}
\label{sec:intro}
%[DONE]

Deep convolutional neural networks (CNNs) with 2D convolutions and small kernels~\cite{vgg}, have achieved state-of-the-art results for several speech recognition tasks~\cite{ibm2015,ibm2016,microsoft2016,vdcnn,vdcnn_adapt}. The accuracy of those models grows with their complexity, leading to redundant latent representations. Several approaches have been proposed in the literature to reduce this redundancy~\cite{pruning1,thinet,condensenet,shufflenet,efficientnet}, and therefore to improve their efficiency. 
%Lower computational cost and memory footprint is a desired feature for real-time decoding and deployment on mobile devices.

Octave convolutional layers~\cite{OctConv} address the problem of spatial redundancy in feature maps by learning feature representations at high and low resolutions. The low resolution processing path increases the size of the receptive field in the original input space, which is a plausible explanation of the improved performance for image classification. We extend the octave convolution concept to multi-scale octave convolutional layers, which include lower resolution feature maps with a higher compression rate (reduction by more than one octave), and the use of more than two feature map tensor groups in order to be learn representations at multiple scales. 
%We hypothesize that multi-scale representation learning is better suited for noisy speech recognition, while further reducing of the computational and memory cost.

%We hypothesize that the decomposition of the output of a convolutional layer into feature maps of different spatial resolutions can also increase the robustness of learned representations, especially in the context of robust speech recognition. 
Multi-scale processing have been previously proposed for a variety of speech recognition tasks~\cite{wavelets1,wavelets2,toth,blnet,multi_span}. In deep CNN acoustic models, some of the feature maps may need to represent information which varies at a lower rate, such as the characteristics of the speaker or background noise,  compared to the information necessary for phonetic discrimination. 
%Therefore,  it  should  be  possible  to  compress some of the feature maps such that they contain information complementary to the feature maps representing fine details differentiating pronunciations.  
Spatial average pooling in a low resolution group of feature maps can be interpreted as a form of low-pass filtering, providing smoothed representations of the observed data, potentially leading to improved performance. 

We investigate the use of multi-scale octave convolutional layers for robust speech recognition, and attempt to shed more light on the explainability of the models by evaluating the robustness of the learned representations using an affine transformation loss to measure the similarity between clean and noisy encodings. 
%Accuracy, efficiency and robustness have not been previously addressed and analyzed jointly in multi-scale networks.
% batchnorm -> relu order

\section{Multi-scale octave convolutions}
\label{sec:method}
%[DONE]

% In this section we present 
% %multi-octave feature representation and Mutli-Octave Convolution. We also describe the method proposed to measure the robustness of inter- and intra-model hidden representations.
% the multi-octave convolutional layer (MultiOctConv) and describe the method proposed to measure the robustness of inter- and intra-model hidden representations.

% \subsection{Multi-octave convolutional layer}
% \label{sec:MultiOctConv}

\begin{figure}[t]
    \centering
    \includegraphics[scale=0.38]{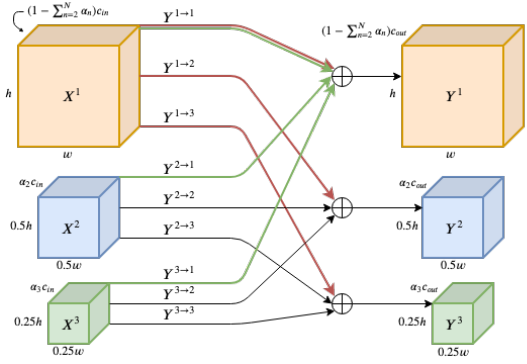}
    \caption{Multi-octave convolution scheme for 3 resolution groups. Red and green arrows show the connections for the initial and final MultiOctConv layers, respectively. $N$ corresponds to the total number of groups in the MultiOctConv layer ($N=3$ in this example). $\alpha_{n}$ is a fraction of channels corresponding to group $n$. $h$ and $w$ are spatial dimensions.}
    \label{fig:octconv}
\end{figure}

An octave convolutional layer~\cite{OctConv} factorizes the output feature maps of a convolutional layer into two groups.
%at spatial frequencies one octave apart, in order to address spatial redundancy in the feature maps. 
The resolution of the low-frequency feature maps is reduced by an octave -- height and width dimensions are divided by $2$. In this work, we explore spatial reduction by up to 3 octaves --  dividing by $2^t$, where $t=1,2,3$ -- and for up to 4 groups.  We refer to such a layer as a multi-octave convolutional  (MultiOctConv) layer, and an example with three groups and reductions of one and two octaves is depicted in Fig.~\ref{fig:octconv}. 

In a vanilla CNN
%convolutional layer, the feature tensors 
the convolutions have the same spatial resolution throughout the network.  An octave convolutional (OctConv) layer is divided into high- and low-frequency feature maps and a multi-octave convolutional (MultiOctConv) layer has feature maps reduced by multiple octaves.  Let the input feature tensor be $X \in \mathbb{R}^{c_{in} \times h \times w}$, where $c_{in}$ denotes the number of input channels and $h$ and $w$ correspond to the spatial dimensions.  In a MultiOctConv layer working at 3 resolutions, $X$ is factorized along the channel dimension into $X = \{X^1, X^2, X^3\}$.  The first tensor group tensor, $X^1$, is a representation at the same spatial scale as $X$. The spatial dimensions of the second and third group tensors, $X^2$ and $X^3$, are reduced by one and two octaves respectively. 

The dimensions of the input tensors $X^1$, $X^2$ and $X^3$ are described in Fig.~\ref{fig:octconv}. The fraction of the channels for each group is denoted with $\alpha_{n} \in [0,1]$, where $\sum_{n=1}^{N} \alpha_{n} = 1$ for $N$ resolution groups in the MultiOctConv layer. For simplicity, we use the same $\alpha_{n}$ for input and output representations within the same scale group. Similarly, the output tensors are also factorized into $Y = \{Y^1, Y^2, Y^3\}$. Their dimensions are analogous to the dimensions of the input tensors and are described in Fig.~\ref{fig:octconv}. To compute $Y^1$, $Y^2$ and $Y^3$ we operate directly on the factorized input tensors $X^1$, $X^2$ and $X^3$. Inter-frequency information update is implemented as a sum of feature maps from different resolution groups. To be able to sum those representations for a desired output scale, the spatial dimensions of the input tensors must be the same. For this reason, two operations are employed: spatial average pooling \texttt{pool($X, p$)} and bilinear interpolation \texttt{upsample($X, u$)}, where $p$ is the kernel size and stride for the the 2D pooling layer and $u$ is the upsampling factor. The output MultiOctConv representations are therefore computed as
\begin{align*}
\begin{split}
    Y_1 = f(X^1; W^{1\rightarrow1}) +  \texttt{upsample}(f(X^2;W^{2\rightarrow1}), 2) + \\
    + \texttt{upsample}(f(X^3;W^{3\rightarrow1}), 4) 
\end{split}
\\ \smallskip
\begin{split}
    Y_2 = f(X^2; W^{2\rightarrow2}) + \texttt{upsample}(f(X^3;W^{3\rightarrow2}), 2) + \\
    + f(\texttt{pool}(X^1, 2); W^{1\rightarrow2})
\end{split}
\\ \smallskip
\begin{split}
    Y_3 = f(X^3; W^{3\rightarrow3}) + f(\texttt{pool}(X^1, 4); W^{1\rightarrow3}) + \\
    + f(\texttt{pool}(X^2, 2); W^{2\rightarrow3})
\end{split}
\end{align*}

\noindent where $f(.)$ is the convolution function and  $W^{n_{in}\rightarrow{n_{out}}}\in\mathbb{R}^{c_{in} \times k \times k \times c_{out}}$ is the convolution filter for a $k \times k$ kernel. We call the information update ``intra-frequency'' when $n_{in} = n_{out}$, and ``inter-frequency'' when $n_{in} \neq n_{out}$. It is important to note that the convolution $f(.)$ operates on the tensors compressed with average pooling and on the tensors before upsampling, making the design more efficient. The number of parameters in the MultiOctConv layer is the same as in a vanilla convolutional layer.

\subsubsection*{Robustness of learned representations}
\label{sec:robustness}

\begin{figure}[t]
    \centering
    \includegraphics[scale=0.4]{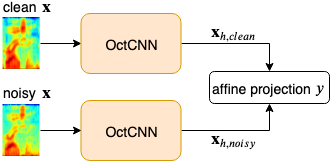}
    \caption{Proposed method to measure the robustness of learned representations.}
    \label{fig:robust}
\end{figure}

To evaluate the robustness of learned representations, we compare the projections of clean and noisy Aurora-4 samples. The similarity between them is measured using the mean squared error (MSE) loss of an affine projection $y$ of $N$ clean to noisy samples (Eq.~\ref{eq:mse}), to take into account permutations of hidden representations and to ensure invariance of the metric to affine transformations of the encodings. The number of units in layer $y$ and the dimensionality $D$ of $\mathbf{x}_{h}$ is 1024.

\begin{equation}
    \theta^* = \argmin_{\theta}{\frac{1}{ND} \sum_{i=1}^N \big\lVert y({\mathbf{x}_{h,clean}^{(i)}, \theta}) - {\mathbf{x}_{h,noisy}^{(i)}} \big\rVert ^2}
    \label{eq:mse}
\end{equation}

We use the Aurora-4 test sets and compare clean encodings $\mathbf{x}_{h,clean}$ with noisy encodings $\mathbf{x}_{h,noise}$, obtained as the activations from the last convolutional layer with a forward pass through a trained model. Both hidden representations were obtained for CNN and octave CNN (OctCNN) models in order to compare representations between the models. Also, for intra-model comparison, we evaluate the loss with the encodings from high and low-resolution groups (paths $Y^{1\rightarrow1}$ and $Y^{2\rightarrow1}$). This analysis aims to evaluate if the low-resolution groups for noisy samples are indeed more similar to the clean ones than the high-resolution encodings, suggesting more robust representations. We optimize the parameters of $y$ with back-propagation using a fixed number of 3 epochs and we report the validation loss for Aurora-4 test sets.

\section{Experimental setup}
\label{sec:setup}
%[DONE]

\noindent \textbf{Aurora-4}~\cite{aurora}: We evaluate our models on the simulated multi-condition Aurora-4 dataset, consisting of $\sim$15h of audio for training and $\sim$9h for testing. The test set is divided into 4 subsets: A, B, C, and D. Subset A contains clean-condition recordings, subset B has 6 noise types added to the recordings (car, babble, restaurant, street, airport, train), subset C is recorded with a mismatched microphone, and subset D is recorded with a mismatched microphone and with noise added. In our experiments, we use multi-condition GMM-HMM forced alignments as targets for CNN training. The number of CD states for Aurora-4 is 3422.

\noindent \textbf{AMI}~\cite{ami}: AMI contains $\sim$100h of meeting recordings, captured by an independent headset microphone (IHM), single distant microphone (SDM), and multiple distant microphones (MDM), where the mics are combined using the BeamformIt~\cite{beamformit} toolkit. We train our models using the MDM data and evaluate the models for all 3 types of recordings to analyze the effect of mismatched training/testing conditions. We use the suggested train/dev/eval data split~\cite{pawel_ami}, and we evaluate the models on both dev and eval sets. The number of CD states for AMI is 3984.

\noindent \textbf{Features}: In our experiments, we use 40-dimension mel-scaled filterbank (FBANK) features with \{-5,..,5\} context for splicing, resulting in a $40\times11$ input feature map. 

\noindent \textbf{Models}: Our baseline CNN model~\cite{my_asru_2017} consists of 15 convolutional and one fully-connected layer. We use $3\times3$ kernels throughout the network. We start with 64 output channels in the first layer and double them after 3 and 9 layers. We use batch normalization in every convolutional layer, and ReLU afterwards (unless a reverse order is noted). The initial learning rate is 0.001. We use early stopping for training.

\iffalse
\begingroup
\renewcommand{\arraystretch}{0.5}
\begin{table}[]
    \centering
    \begin{tabular}{l|r|r|r|r}
        layer (L) & $c_{in}$ & $d_{in}$ & $c_{out}$ & $d_{out}$ \\ \hline
        Conv1 &	1 & 40x11 & 64 & 40x11 \\
        Conv2 &	64 & 40x11 & 64	& 40x11 \\
        Conv3 & 64 & 40x11 & 64 & 20x11 \\
        Conv4 & 64 & 20x11 & 128 & 20x11 \\        
        Conv5 & 128 & 20x11 & 128 & 20x11 \\
        Conv6 & 128 & 20x11 & 128 & 10x11 \\
        Conv7 & 128 & 10x11 & 128 & 10x11 \\
        Conv8 & 128 & 10x11 & 128 & 10x11 \\
        Conv9 & 128 & 10x11 & 128 & 5x6 \\
        Conv10 & 128 & 5x6 & 256 & 5x6 \\
        Conv11 & 256 & 5x6 & 256 & 5x6 \\
        Conv12 & 256 & 5x6 & 256 & 3x3 \\
        Conv13 & 256 & 3x3 & 256 & 3x3 \\
        Conv14 & 256 & 3x3 & 256 & 3x3 \\
        Conv15 & 256 & 3x3 & 256 & 2x2 \\
        FC 	& 256 & 2x2 & 3422 & 1x1
    \end{tabular}
    \caption{Baseline CNN network structure.}
    \label{tab:vdcnn}
\end{table}
\endgroup
\fi

\section{Results}
\label{sec:results}
%[DONE]

We present our results in terms of accuracy and robustness on Aurora-4 and AMI, as well as in terms of the computational cost, which is calculated as the number of multiply-accumulate operations (MACCs) performed for a single input feature map. The cost reduction 
%for OctCNNs and MultiOctCNNs
when using octave convolutions stems from reduced dimensions $c_{in}$, $c_{out}$, $h$, and $w$ compared to a vanilla convolutional layer. 
% write about reduced memory cost?

% \subsection{Aurora-4}
% \label{sec:results_aurora4}
\smallskip
\noindent
\textbf{Aurora-4:}
Results for Aurora-4 are presented in Table~\ref{tab:aurora_wer}. We replace vanilla convolutional layers of our baseline model (CNN) with OctConv and MultiOctConv layers. 
%OctCNN is a model with 2 groups, one octave apart. MultiOctCNN is a model with more groups or / and more octaves apart. 
We first evaluate which layers can be replaced and find that all but the first layer, operating directly on the input representation, should be replaced for the best performance. This approach (L2-L15) is also the least costly.
% Conv2 is now an initial layer (with only the red paths), so it has much less connections than when Conv2 is not an initial layer. For Conv1, the initial layer effect is not that pronounced because there is only 1 input feature map.
Reducing the ratio of low-resolution representations to 0.125 improves the WER for the mismatched microphone scenario C, but not for all test conditions. Applying batch normalization after ReLU is beneficial for test set C and D. For OctCNN models, the WER for test set D dropped by $\sim0.4\%$ with a compression by one octave, and by another $\sim0.4\%$ with a reversed batch normalization and ReLU order.

\begin{table*}[ht]
    \centering \small
    \begin{tabular}{llc|ccc|c|rrrr|r}
         Model & OctConv & $\alpha$ (low $\rightarrow$ high) & $2^1$ & $2^2$ & $2^3$ & \#MACCs (M) & A & B & C & D & Avg. \\ \hline
         CNN & - & [0, 1] & - & - & - & 174.7 & 2.19 & 4.68 & 4.22 & 14.53 & 8.69 \\ \hline %vdcnn_allconv_oct_bn_bigger_multi_groups_1_bn_activ_order_all
         OctCNN & L1-L3 & [0.2, 0.8] & $\checkmark$ & - & - & 167.6 & 2.19 & 4.74 & 4.32 & 14.83 & 8.85 \\ %vdcnn_allconv_oct_bn_bigger_multi_groups_2_just_input_bn_activ_order_all
         OctCNN & L1-L15 & [0.2, 0.8] & $\checkmark$ & - & - & 126.9 & 2.22 & 4.61 & 4.30 & 14.40 & 8.61 \\ %vdcnn_allconv_oct_bn_bigger_multi_groups_2_bn_activ_order_all
         OctCNN & L2-L15 & [0.2, 0.8] & $\checkmark$ & - & - & 126.2 & 2.02 & 4.65 & 4.35 & 14.16 & 8.52 \\ %vdcnn_allconv_oct_bn_bigger_multi_groups_2_no_input_bn_activ_order_all
         OctCNN $^\dagger$ & L2-L15 & [0.2, 0.8] & $\checkmark$ & - & - & 126.2 & 2.22 & 4.82 & 4.22 & 13.72 & 8.41 \\ %vdcnn_allconv_oct_bn_bigger_multi_groups_2_activ_bn_no_input 
         OctCNN & L2-L15 & [0.125, 0.875] & $\checkmark$ & - & - & 143.1 & 2.11 & 4.56 & 4.07 & 14.55 & 8.63 \\ \hline %vdcnn_allconv_oct_bn_bigger_multi_groups_2_bn_activ_0125_no_input
         MultiOctCNN & L2-L15 & [0.1, 0.1, 0.8] & $\checkmark$ & $\checkmark$ & - & 120.6 & \textbf{1.98} & 4.51 & 4.11 & 14.00 & 8.37 \\ %vdcnn_allconv_oct_bn_bigger_multi_groups_3_bn_activ_0125_no_input
         MultiOctCNN & L2-L15 & [0.1, 0.1, 0.8] & $\checkmark$ & - & $\checkmark$ & 119.5 & 2.02 & 4.59 & \textbf{3.92} & 13.82 & \textbf{8.31} \\ %vdcnn_allconv_oct_bn_bigger_multi_groups_3_bn_activ_0125_no_input_1_3
         MultiOctCNN $^\dagger$ & L2-L15 & [0.1, 0.1, 0.8] & $\checkmark$ & - & $\checkmark$ & 119.5 & 2.28 & 4.81 & 4.04 & 13.76 & 8.41 \\ %vdcnn_allconv_oct_bn_bigger_multi_groups_3_activ_bn_0125_no_input_1_3
         MultiOctCNN & L2-L15 & [0.1, 0.1, 0.1, 0.7] & $\checkmark$ & $\checkmark$ & $\checkmark$ & \textbf{94.3} & 2.30 & 4.88 & 4.18 & 14.06 & 8.58 \\ %vdcnn_allconv_oct_bn_bigger_multi_groups_4_bn_activ_0125_no_input
         MultiOctCNN & L2-L15 & [0.2, 0.8] & - & $\checkmark$ & - & 115.7 & 2.15 & 4.77 & 4.07 &  13.77 & 8.39 \\ %vdcnn_allconv_oct_bn_bigger_multi_groups_2_bn_activ_all_2_no_input_lf4
         %MultiOctCNN $^\dagger$ & L2-L15 & [0.2, 0.8] & - & $\checkmark$ & - & & 2.17 & 4.80 & 4.56 & 14.00 & 8.54 \\ %vdcnn_allconv_oct_bn_bigger_multi_groups_2_activ_bn_no_input_lf4
         MultiOctCNN & L2-L15 & [0.125, 0.875] & - & $\checkmark$ & - & 136.3 & 2.09 & 4.56 & 4.22 & 14.32 & 8.54 \\ %vdcnn_allconv_oct_bn_bigger_multi_groups_2_bn_activ_all_0125_no_input_lf4
         MultiOctCNN & L2-L15 & [0.2, 0.8] & - & - & $\checkmark$ & 113.5 & 2.09 & 4.54 & 3.94 & 14.03 & 8.39 \\ %vdcnn_allconv_oct_bn_bigger_multi_groups_2_bn_activ_all_02_no_input_lf8
         MultiOctCNN & L2-L15 & [0.125, 0.875] & - & - & $\checkmark$ & 134.9 &2.02 & \textbf{4.50} & 4.17 & 13.87 & 8.32 \\ %vdcnn_allconv_oct_bn_bigger_multi_groups_2_bn_activ_all_0125_no_input_lf8
         MultiOctCNN $^\dagger$ & L2-L15 & [0.125, 0.875] & - & - & $\checkmark$ & 134.9 & 2.32 & 4.73 & 4.24 & \textbf{13.57} & \textbf{8.31} %vdcnn_allconv_oct_bn_bigger_multi_groups_2_activ_bn_0125_no_input_lf8
    \end{tabular}
    \caption{WERs [\%] for Aurora-4 test sets A, B, C and D for octave and multi-octave CNNs. "OctConv" column indicates where a Conv layer was replaced with an OctConv or MultiOctConv. $2^1$, $2^2$ and $2^3$ correspond to the factor of spatial dimension reduction. Models with batch normalization after ReLU are denoted by $^\dagger$.}
    \label{tab:aurora_wer}
\end{table*}
% bn-activ order

The biggest differences between the MultiOctCNN models can be observed for test set D. The models with the lowest WERs are the ones with a spatial reduction by 2 or 3 octaves, and with 2 or 3 groups. This indicates that multi-scale octave convolutions seem to be an effective, as well as an efficient design for processing speech with background noise and channel mismatch. For MultiOctCNNs, batch normalization after ReLU also gives a performance boost for test set D, with a drop to $13.57\%$.

\begin{figure}[t]
    \centering
    \includegraphics[scale=0.17]{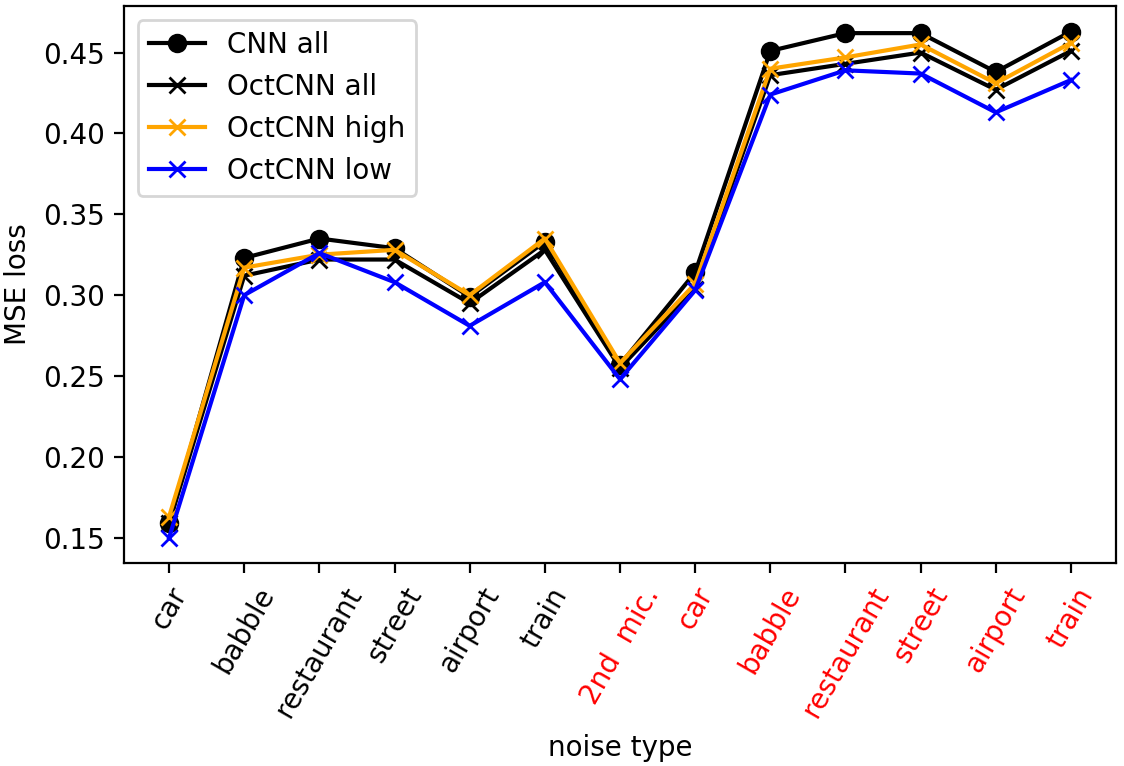}
    \caption{MSE affine transformation loss to measure the similarity of "clean" and "noisy" encodings ($\mathbf{x}_{h,clean}$ and $\mathbf{x}_{h,noisy}$). "all" corresponds to the output of the last convolutional layer (Conv15), "high" and "low" correspond to its $Y^{1\rightarrow1}$ and $Y^{2\rightarrow1}$ branch, respectively.}
    \label{fig:mse}
\end{figure}

To further evaluate the robustness of the latent representations 
%to better understand the inner workings of our models 
we measured the MSE between the (projected) representations,  described above
(Fig.~\ref{fig:mse}). The loss for the activations at the output of Conv15 ("all") is similar for CNN and OctCNN models for test sets B and C, but lower for test set D for OctCNN, indicating that the learned representations are more robust, contributing to lower WERs. As expected, within-model comparison of the loss show that the representations at low resolution are more similar to the clean encodings from test set A than the ones at high resolution. We believe that this effect improves the robustness of latent representations and results in a decreased WER.

\begin{table*}[h!t]
    \centering \small
    \begin{tabular}{llc|ccc|c|rrrrrr}
         &&&&&& & \multicolumn{2}{c}{IHM} & \multicolumn{2}{c}{SDM} & \multicolumn{2}{c}{MDM} \\ 
         Model & OctConv & $\alpha$ (low $\rightarrow$ high) & $2^1$ & $2^2$ & $2^3$ & \#MACCs (M) & dev & eval & dev & eval & dev & eval \\ \hline
         CNN & - & [0, 1] & - & - & - & 175.2 & 33.4 & 38.3 & 49.1 & 54.0 & 43.9 & 48.0  \\ \hline %vdcnn_allconv_oct_bn_bigger_multi_groups_1_bn_activ_order_all_mdm
         OctCNN & L1-L3 & [0.2, 0.8] & $\checkmark$ & - & - & 168.2 & 33.0 & 38.1 & 49.0 & 54.1 & 43.8 & 47.9 \\ %vdcnn_allconv_oct_bn_bigger_multi_groups_2_just_input_bn_activ_order_all_1D_mdm
         OctCNN & L2-L15 & [0.2, 0.8] & $\checkmark$ & - & - & 126.7 & 33.0 & 37.7 & 48.9 & 54.0 & 43.7 & 47.7 \\ %vdcnn_allconv_oct_bn_bigger_multi_groups_2_no_input_bn_activ_order_all_mdm
         OctCNN & L1-L15 & [0.2, 0.8] & $\checkmark$ & - & - & 127.5 & \textbf{32.2} & \textbf{37.2} & 48.3 & 53.5 & 43.1 & 47.3 \\  %vdcnn_allconv_oct_bn_bigger_multi_groups_2_bn_activ_order_all_mdm
         OctCNN & L1-L15 & [0.125, 0.875] & $\checkmark$ & - & - & 144.1 & 32.5 & 37.4 & \textbf{48.2} & \textbf{53.3} & \textbf{42.9} & \textbf{47.2} \\ %vdcnn_allconv_oct_bn_bigger_multi_groups_2_bn_activ_order_all_0125_mdm
         OctCNN$^\dagger$ & L1-L15 & [0.125, 0.875] & $\checkmark$ & - & - & 144.1 & 33.2 & 38.3 & 48.8 & 54.3 & 43.7 & 48.0 \\ \hline %vdcnn_allconv_oct_bn_bigger_multi_groups_2_activ_bn_order_all_0125_mdm
         MultiOctCNN & L1-L15 & [0.1, 0.1, 0.8] & $\checkmark$ & $\checkmark$ & - & 121.6 & 32.8 & 38.1 & 48.9 & 53.9 & 43.7 & 47.9 \\ %vdcnn_allconv_oct_bn_bigger_multi_groups_3_bn_activ_order_all_0125_mdm
         MultiOctCNN & L1-L15 & [0.1, 0.1, 0.8] & $\checkmark$ & - & $\checkmark$ & 120.4 & 33.3 & 38.5 & 49.2 & 54.5 & 44.1 & 48.4 \\ %vdcnn_allconv_oct_bn_bigger_multi_groups_3_bn_activ_order_all_0125_mdm_1_3
         MultiOctCNN & L1-L15 & [0.1, 0.1, 0.1, 0.7] & $\checkmark$ & $\checkmark$ & $\checkmark$ & \textbf{95.2} & 33.7 & 38.7 & 49.5 & 54.6 & 44.1 & 48.4 \\ %vdcnn_allconv_oct_bn_bigger_multi_groups_4_bn_activ_order_all_0125_mdm
         MultiOctCNN & L1-L15 & [0.125, 0.875] & - & $\checkmark$ & - & 136.9 & 33.6 & 38.6 & 49.7 & 54.6 & 44.3 & 48.4 \\ %vdcnn_allconv_oct_bn_bigger_multi_groups_2_bn_activ_order_all_0125_mdm_lf4
         MultiOctCNN & L1-L15 & [0.125, 0.875] & - & - & $\checkmark$ & 135.4 & 32.9 & 38.1 & 49.1 & 54.3 & 43.8 & 48.0 \\ %vdcnn_allconv_oct_bn_bigger_multi_groups_2_bn_activ_order_all_0125_mdm_lf8
    \end{tabular}
    \caption{WERs [\%] for models trained on AMI MDM and evaluated on IHM, SDM and MDM conditions. "OctConv" column indicates where a Conv layer was replaced with an OctConv or MultiOctConv. $2^1$, $2^2$ and $2^3$ correspond to the factor of spatial dimension reduction.  Models with batch normalization after ReLU are denoted by $^\dagger$.}
    \label{tab:ami_wer}
\end{table*}

% \subsection{AMI}
% \label{sec:results_ami}
\smallskip
\noindent
\textbf{AMI:}
Results for AMI are presented in Table~\ref{tab:ami_wer}. In contrast to the Aurora-4 findings, better performance was achieved with an all OctCNN model (L1-L15). This is an interesting finding, and we believe that the multi-scale processing of the input feature space is beneficial for AMI MDM because of the reverberation in the data. The reverbarated input time$\times$freq representation can be viewed as a spatially redundant one, therefore the OctConv layer applied to the input representation is effective. Unfortunately, the only MultiOctConv model superior to the baseline CNN is the one with 3 groups with a spatial reduction by 1 and 2 octaves. This result indicates that the spatial redundancy for this architecture for AMI MDM is not degrading the performance. However, in terms of the computational cost, we can reduce the \#MACCs by a factor of 1.8 with only a small WER increase for a model with 4 resolution groups.

\section{Conclusions}
\label{sec:conclusions}
%[DONE]

We have presented multi-scale octave CNN models for robust and efficient speech recognition. We build on Chen et al~\cite{OctConv}, applying the method to robust ASR and extending it to multiple resolution groups with a spatial reduction of more than one octave. 
%We hypothesize that OctCNNs and multi-scale OctCNNs not only reduce spatial redundancy in the feature maps but also improve the robustness of learned representations. 
Our experiments confirm that multi-scale processing of the hidden representations is not only more computationally efficient, but also improves the recognition. Similarity measures between clean and noisy encodings  indicates that multi-scale processing in a deep CNN acoustic model improves the robustness of learned representations, especially in the additive noise and mismatched microphone scenario. The gain of the octave convolutions was also observed for AMI MDM data with significant reverberation, when applied to the input feature space. However, the model performance for AMI MDM was not improved with multi-octave convolutions. More careful tuning of the $\alpha$ hyperparameter could improve the results, as it %defines model capacity and robustness trade-off. Optimizing $\alpha$, which 
controls the ratio of multi-scale feature maps in the model, enabling both learning of fine-grained representations preserving the details necessary for  phonetic discrimination, as well as smoothed, more invariant representations improving the robustness of the model. It would also be possible to set  $\alpha$ layer-by-layer to enable the fractions of channels at different resolutions to vary according to the depth of the representation. 

We proposed a single projection layer MSE loss to measure the affine relationship of clean and noisy hidden representations. With this approach, we evaluated the robustness of the encodings and improved the explainability of our models. More thorough analysis of the representations learned is an interesting future direction. We confirmed that the noisy lower resolution representations are more similar to the clean counterparts than high resolution representations, and thus are more robust. However, we did not investigate the reason for the increased similarity, leaving future work to ascertain if the lower resolution group corresponds to better speaker or noise characteristics, or more invariant phonetic representations.

%Further work:
% evaluate with pertinent positives

\bibliographystyle{IEEEbib}
\bibliography{refs}
%[DONE]

\end{document}